\documentclass[10pt,twocolumn,letterpaper]{article}
\pdfoutput=1

\usepackage{iccv}
\usepackage{times}
\usepackage{epsfig}
\usepackage{graphicx}
\usepackage{amsmath}
\usepackage{amssymb}

\usepackage{booktabs}
\usepackage{makecell}
\usepackage{amsfonts}
\usepackage[table]{xcolor}
\usepackage{subcaption}
\usepackage{multirow}
\usepackage{colortbl}

\usepackage{algorithm}
\usepackage{algorithmic}

\usepackage[pagebackref=true,breaklinks=true,letterpaper=true,colorlinks,bookmarks=false]{hyperref}

\iccvfinalcopy 

\begin{document}

\title{Proposal-Level Unsupervised Domain Adaptation for Open World Unbiased Detector}
\author{\textbf{Xuanyi Liu}$^1$,~~~~~\textbf{Zhongqi Yue}$^{1,2}$,~~~~\textbf{Xian-Sheng Hua}$^{3}$\\
{\small $^{1}$Nanyang Technological University,~~~~$^{2}$Damo Academy, Alibaba Group,~~~~~$^{3}$Terminus Group} \\
{\tt\small xuanyi001@e.ntu.edu.sg,~yuez003@e.ntu.edu.sg,~huaxiansheng@gmail.com}
%
}

\maketitle

\begin{abstract}
Open World Object Detection (OWOD) combines open-set object detection with incremental learning capabilities to handle the challenge of the open and dynamic visual world. Existing works assume that a foreground predictor trained on the seen categories can be directly transferred to identify the unseen categories' locations by selecting the top-$k$ most confident foreground predictions. However, the assumption is hardly valid in practice. This is because the predictor is inevitably biased to the known categories, and fails under the shift in the appearance of the unseen categories. In this work, we aim to build an unbiased foreground predictor by re-formulating the task under Unsupervised Domain Adaptation, where the current biased predictor helps form the domains: the seen object locations and confident background locations as the source domain, and the rest ambiguous ones as the target domain. Then, we adopt the simple and effective self-training method to learn a predictor based on the domain-invariant foreground features, hence achieving unbiased prediction robust to the shift in appearance between the seen and unseen categories. Our approach's pipeline can adapt to various detection frameworks and UDA methods, empirically validated by OWOD evaluation, where we achieve state-of-the-art performance. Codes are available at \url{https://github.com/lxycopper/PLU}
\end{abstract}.
\begin{figure}[t]
  \centering
  \subcaptionbox{Closed-Set Object Detection}{\includegraphics[width=0.95\linewidth]{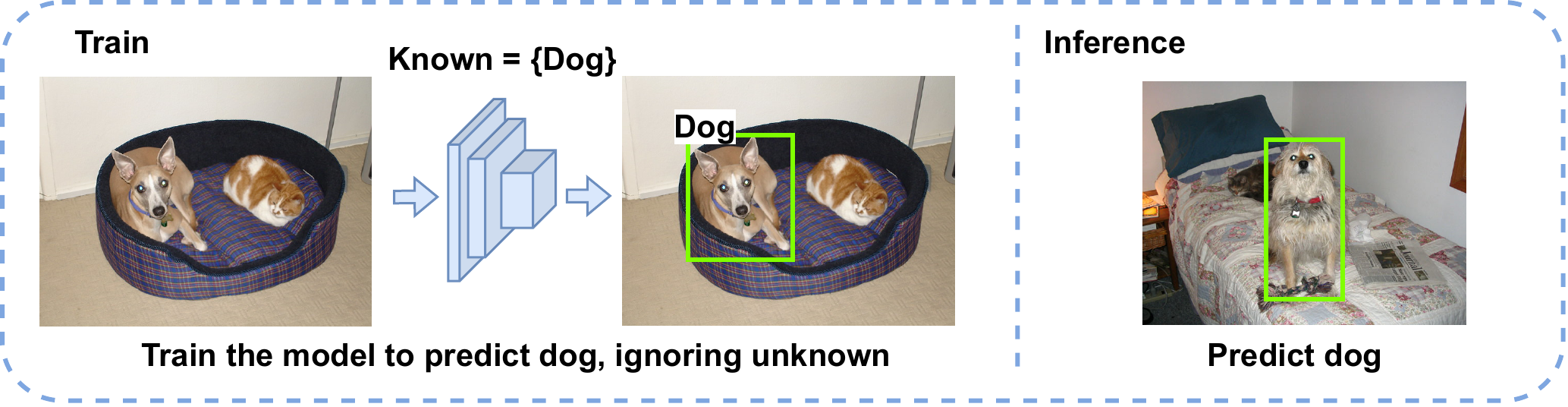}}
\subcaptionbox{Open World Object Detection}
    {\includegraphics[width=0.95\linewidth]{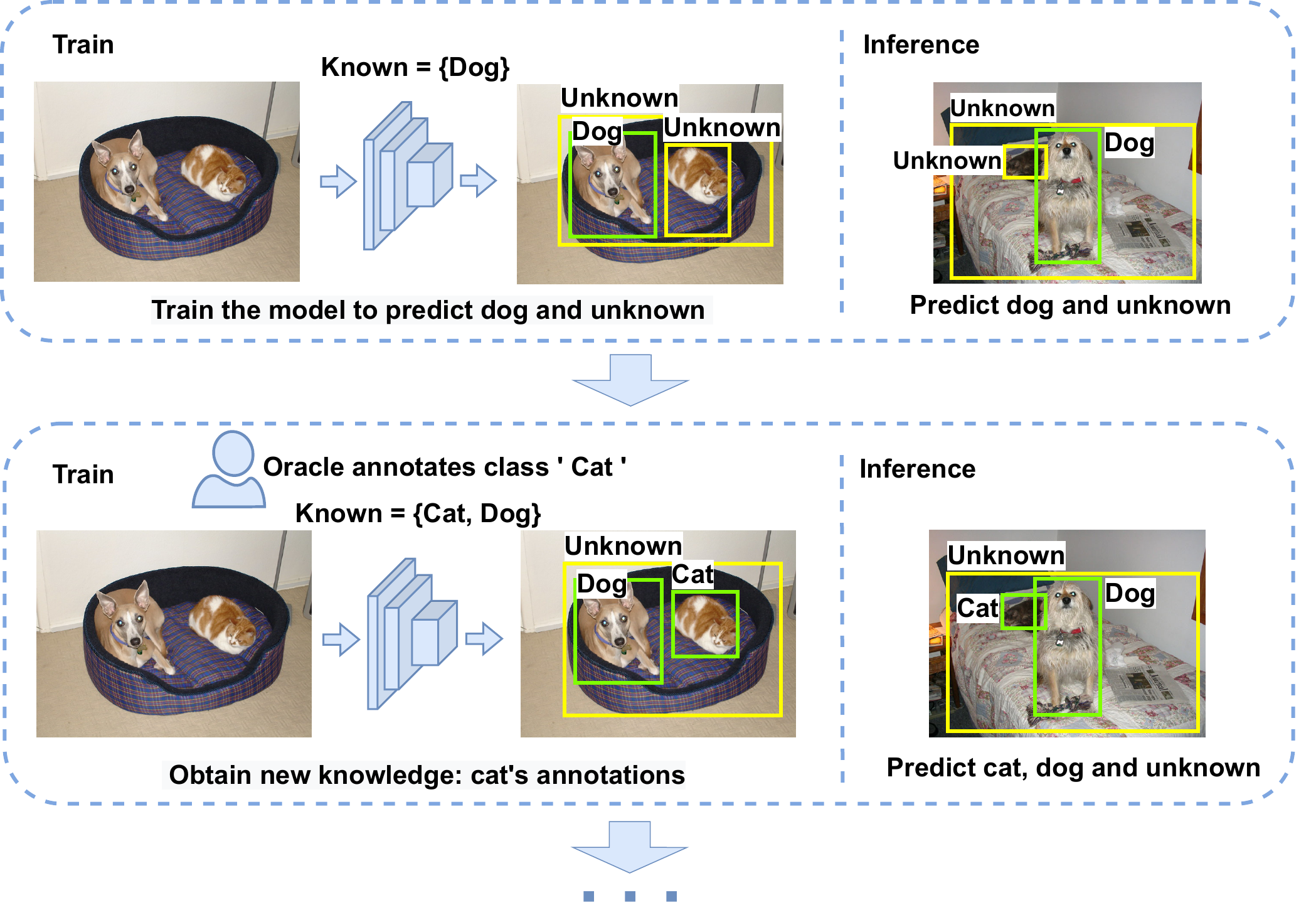}}
   \caption{(a) Traditional closed-set object detector trained on dog classes can only detect dogs, failing to recognize other unknown classes. (b) Open world object detector can predict dog and unknown classes. After an oracle offers more knowledge, such as annotating some of the unknown classes as 'cat', the detector can incrementally learn to predict more $k$ known classes (dog, cat) and unknown classes. }
   \label{fig:owod problem}
\end{figure}

\section{Introduction}
\label{sec:1}


An object detector deployed in real life is constantly challenged by the vast open and dynamic visual world, where any scene may contain objects unseen by the object detector in training, requiring the detector to recognize new categories. For example, it is important for an autonomous driving system to mark unfamiliar objects on roads as `unknown' to take necessary precautions or to incrementally learn a newly implemented road sign. As shown in Figure~\ref{fig:owod problem}a, the existing object detection task~\cite{ren2015faster,carion2020end,zhu2020deformable} under the conventional close-world paradigm are far from achieving this---they can recklessly predict unseen as one of the seen categories and require expensive model re-training to add new object categories. The challenging needs posed by the open-world call for a new paradigm, which is known as the Open-World Object Detection (OWOD)~\cite{joseph2021towards}.


\begin{figure*}[t]
  \centering
  \subcaptionbox{ 
   top-$k$
  selection strategy on unmatched proposals.}{\includegraphics[width=0.8\linewidth]{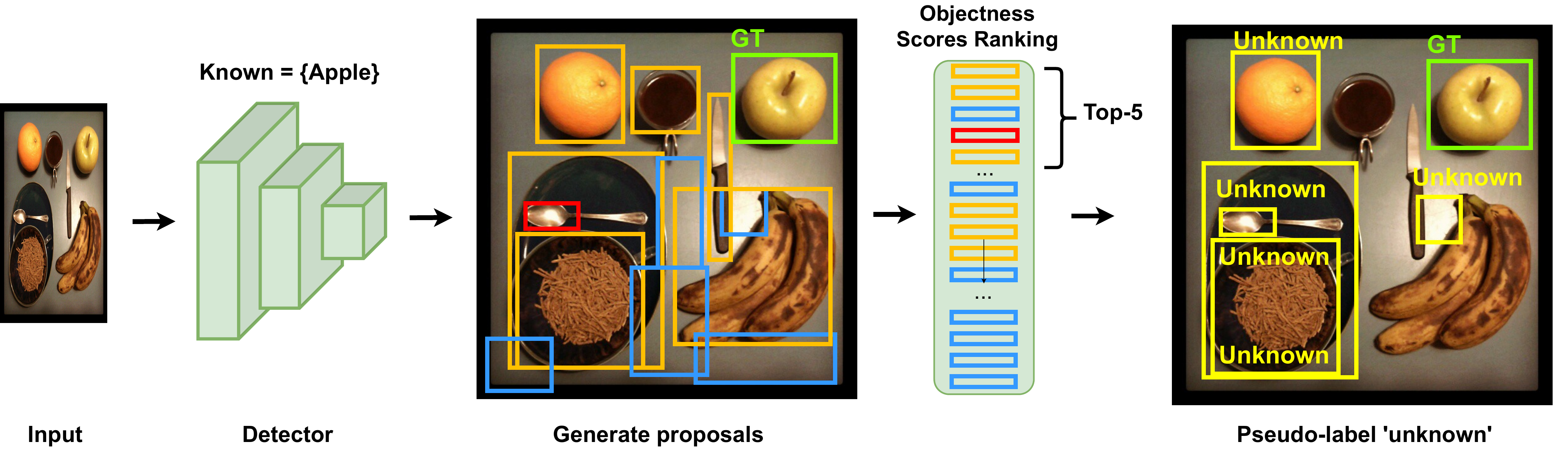}}
  \subcaptionbox{detector is biased.}
    {\includegraphics[width=0.29\linewidth]{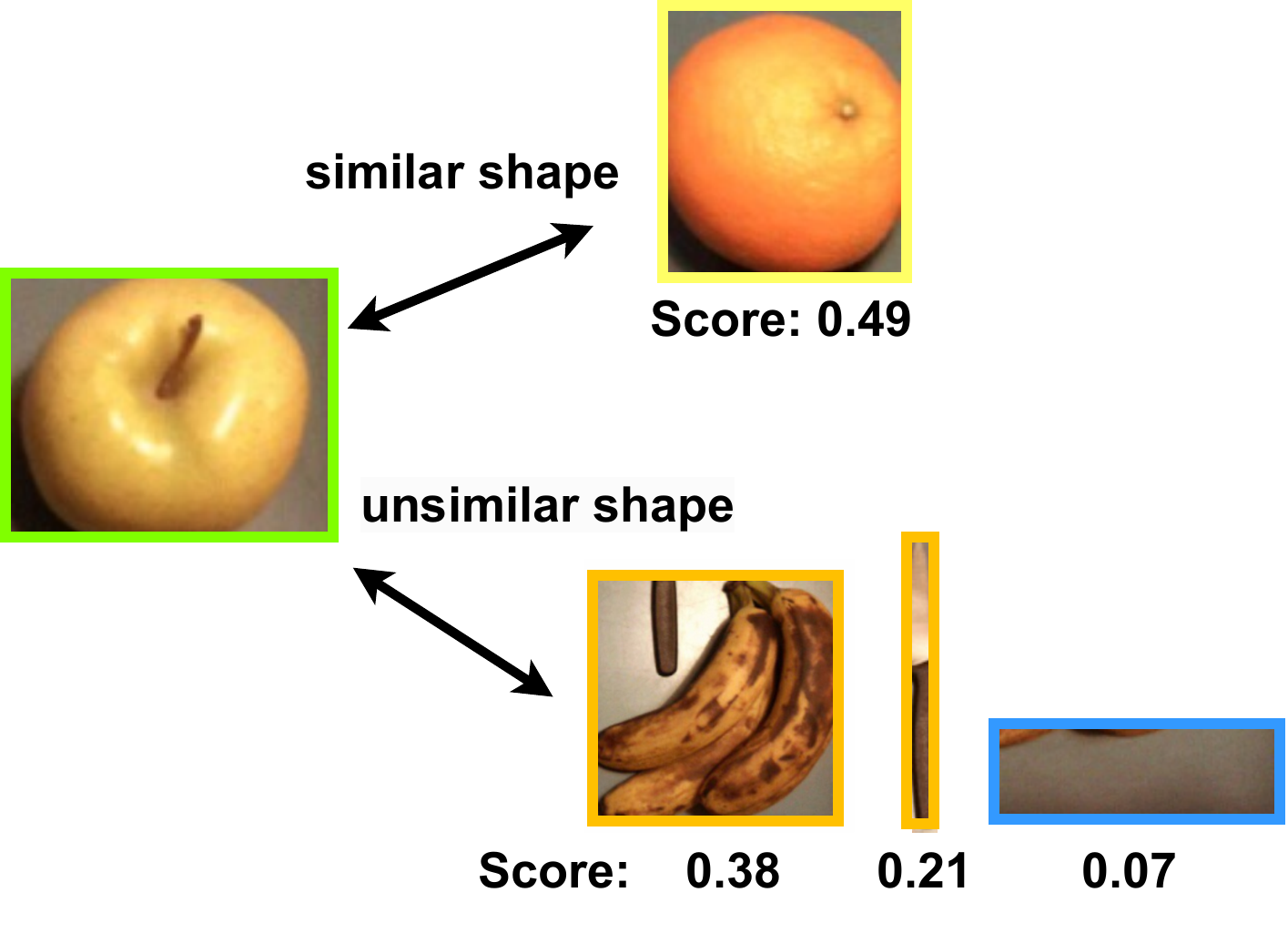}}
    \subcaptionbox{proposals sorted by objectness scores.}
    {\includegraphics[width=0.7\linewidth]{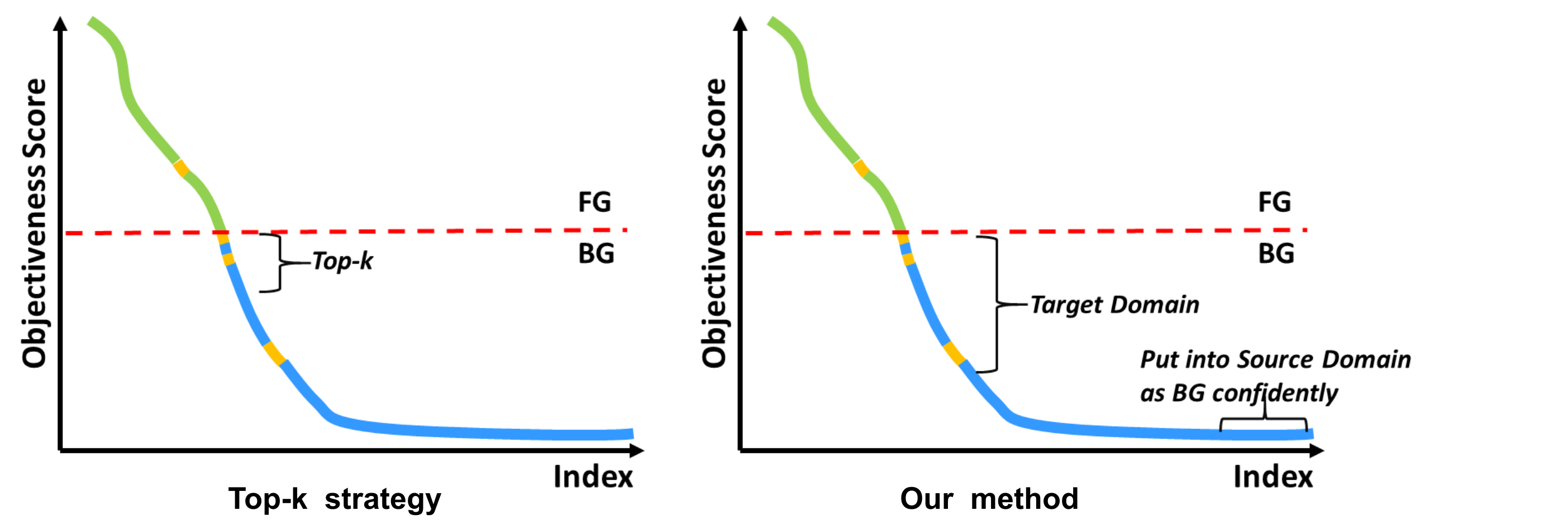}}

   \caption{
   (a) shows a general top\text-$k$ selection strategy. The detector will compute an objectness score for each proposal. The proposals that have small overlaps with ground-truth (green boxes) will be ranked by the objectness scores and the top-$k$ ones will be pseudo-labeled `unknown'. It can be observed that top-$k$ is not flexible. The $k$ proposals fail to contain all unknown objects (orange boxes and red boxes) and mistakenly contain background proposals (blue boxes). (b) shows the detector's bias when computing objectness scores. The detector tends to give higher scores to unknown objects who have a similar appearance to the known classes (apple). The orange's score is much higher than the banana and knife, while the background has a much lower score. (c) shows that the top-$k$ strategy neither guarantees that the $k$ proposals are all unknown objects nor covers all unknown proposals. It also shows that the proposals with the lowest objectness scores are always BG. }
   \label{fig:2}
\end{figure*}

OWOD can be broadly described as open-set object detection with incremental learning capabilities. OWOD detector can incrementally learn and recognize known and unknown classes step-by-step, starting with the known classes and gradually adding the unknown ones. As illustrated in Figure~\ref{fig:owod problem}b, the objects ignored by traditional closed-set detectors (e.g.\, cat, bed) which are trained on known classes (e.g.\, dog) will be predicted as `unknown' by OWOD predictor. Once unknown categories are labelled, the model updates incrementally without training from scratch.

Naturally, detecting unknown objects is a crucial step in OWOD, which serves as a foundation for the subsequent steps. Unfortunately, it is challenging to implement because there are no labels within unknown classes used as supervision signals. As shown in Figure~\ref{fig:2}a, to address the problem, existing methods~\cite{joseph2021towards,gupta2022ow,zhao2022revisiting} generally adopt a top-$k$ selection strategy for the pseudo-label generation. Specifically, in this strategy, each unmatched proposal, which is unable to match with ground truths due to their small overlap (e.g.\ IoU$<$0.5), is assigned an objectness score. The score is computed from the detector's extracted features, \textbf{and} a higher score reflects that the proposal is more likely to contain an object, then $k$ proposals with the highest scores are annotated an unknown class pseudo label.

However, the $top\text{-}k$ selection strategy has two drawbacks: 1) \textbf{not flexible}. $k$ is a fixed hyper-parameter that requires careful manual tuning. Moreover, as there is significant variation between images, a fixed hyper-parameter may not be effective in handling all situations. 2) \textbf{biased}. The objectness scores are computed by the detector trained with known class annotations so that the predictor is inevitably biased to the known categories. That is to say, the detector has the tendency to assign higher scores to objects similar to known classes as an example shown in Figure~\ref{fig:2}b.




To address this problem, we attempt to build an unbiased foreground/background (FG/BG for brevity) predictor to replace the $top\text{-}k$ selection strategy, assigning unknown pseudo-labels to unmatched proposals. The predictor should be robust to the shift in appearance from the known to unknown categories. Specifically, we obtain inspiration from the well-studied Unsupervised Domain Adaptation (UDA)~\cite{ganin2015unsupervised,tsai2018learning,sohn2020fixmatch} to reformulate OWOD task. UDA aims to learn a model in a supervised source domain that generalizes to an unsupervised target domain with a significant domain shift. We illustrate how to reformulate UDA for OWOD as following:


\noindent\textbf{Source Domain (Known Classes)}.
As known classes proposals annotations are provided, we assign them a label of 1 (i.e., FG).
Furthermore, we empirically discover in Figure~\ref{fig:2}c that the unmatched proposals with lowest objectness scores are confidently BGs. Hence we assign the corresponding proposals the label 0 (i.e., BG).

\noindent\textbf{Target Domain (Unkown Classes)}. All the rest unmatched region proposals form the target domain. We do not label them as biased predictions are often ambiguous on them. We use the self-training method of UDA, whose empirical and theoretical results show its effectiveness~\cite{sakaridis2018model,yang2020fda,zou2018unsupervised,zou2019confidence}. Overall, we term our approach \textbf{P}roposal-\textbf{L}evel \textbf{U}nsupervised Domain Adaptation (\textbf{PLU}). 

Our contributions include:

$\bullet$ To the best of our knowledge, our method is the first attempt to introduce the idea of UDA into OWOD. The UDA works on the object proposals level,  which will give insights to the community.

$\bullet$ We develop a simple UDA module help to select unmatched proposals to pseudo-label them as unknown and BG. And we propose a pipeline to extend the UDA module to various object detection frameworks.

$\bullet$ Our extensive experiments based on Faster-RCNN framework and DDETR-framework demonstrate the effectiveness of the UDA Module. In OWOD tasks, the framework with our UDA module have gained state-of-the-art performance.

\section{Related Work}
\label{sec:2}

\noindent\textbf{Open World Object Detection.}
The formulation of OWOD is firstly proposed by
Joesph et.al~\cite{joseph2021towards}, and they also propose a Faster-RCNN~\cite{ren2015faster} based approach termed as ORE and related evaluation protocal. Zhao et.al~\cite{zhao2022revisiting} follow the Faster-RCNN structure and add an auxilary proposal advisor to help identify unknown proposals. Wu~\cite{wu2022uc} modify ORE and propose Unknown Classified OWOD (UC-OWOD) problem, an extension of OWOD the UC-OWOD, which classifies the unknown instances into different categories. 
Gupta et.al~\cite{gupta2022ow} propose the first OWOD approach based on DETR~\cite{carion2020end} using the attention activation maps to pseudo-label unknown classes. Maaz et.al~\cite{maaz2022class} has proposed a class-agnostic object detection method with a multi-modal vision transformer, which can be adapted in OWOD by modifying the input prompt. All the previous work is relied on one kind of specific object detection framework.

\noindent\textbf{Unsupervised Domain Adaptation.}
Unsupervised domain adaptation (UDA) is proposed as a viable solution to migrate knowledge learned from a labeled source domain to unlabeled target domains~\cite{ganin2015unsupervised,kouw2018introduction}. The solution to UDA is primarily classified into self-training and adversarial learning~\cite{liu2022deep, hoyer2022daformer}. Adversarial training methods aim to align the distributions of source and target domain at input \cite{gong2019dlow,hoffman2018cycada}, feature~\cite{hoffman2016fcns,tsai2018learning,hoyer2022daformer}, output ~\cite{tsai2018learning,vu2019advent}, or patch
level~\cite{tsai2019domain} in a GAN manner~\cite{ganin2016domain,goodfellow2020generative}. In self-training, the target domain is labelled by pseudo-labels~\cite{lee2013pseudo}. Pseudo-labels can be pre-computed in offline mode then be used to train the model~\cite{sakaridis2018model,yang2020fda,zou2018unsupervised,zou2019confidence}. Also, pseudo-labels can be generated online during the training. For the transferring performance's concern, pseudo-label prototypes ~\cite{zhang2021prototypical} or consistency regularization~\cite{sohn2020fixmatch,tarvainen2017mean} based on data augmentation~\cite{araslanov2021self,choi2019self,melas2021pixmatch} or domain-mixup~\cite{tranheden2021dacs,zhou2022context} are used.

\noindent\textbf{Unsupervised Domain Adaptation for Object Detection.}
There are a lot of work utilizing UDA in object detection to mitigate the gap between the source domain and the target domain.
Adversarial feature learning methods~\cite{zhao2020adaptive,su2020adapting,vs2021mega,zhang2021detr,chen2021scale} and self-training~\cite{roychowdhury2019automatic,rodriguez2019domain,khodabandeh2019robust,zhao2020collaborative,soviany2021curriculum,wang2023fvp} methods are proposed. However, these methods are based on transferring the invariant features between two domains with a data distribution shift at image-level. Wu et. al~\cite{wu2021instance} and Wang et.al~\cite{wang2021robust} have proposed domain adaptation approaches to transfer features at instance-level. But their work aims to align features between instances from images in different domains not within images and their objective is not to help locate instances belonging to novel classes. There have not been previous UDA work for Open World Object Detection.

\section{Preliminaries}
\label{sec:3}

\label{sec:3.1}


\begin{figure*}
    \centering
    \includegraphics[width = 1.0\linewidth]{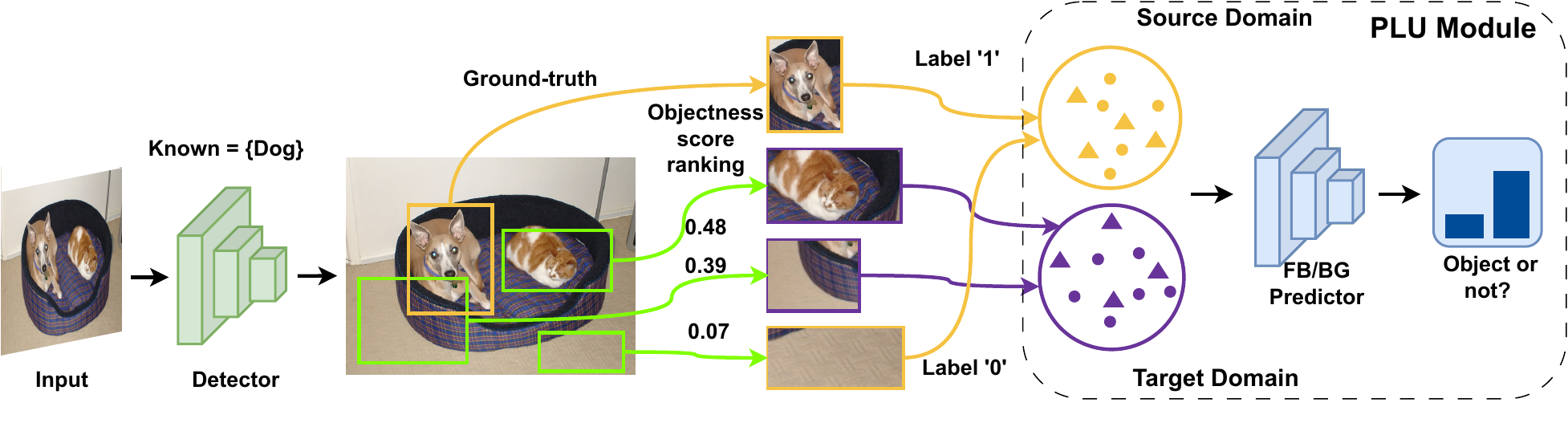}
    \caption{The detector trained with dogs will produce a lot of unmatched proposals (green boxes) with objectness scores. After ranking the unmatched proposals, the lowest proposals labeled as `0' and the ground-truth dog annotation (yellow boxes) form the Source Domain, while the rest of the unmatched proposals (purple boxes) form the Target Domain. These domains will help train the FG/BG predictor together.}
    \label{fig:3}
\end{figure*}

\noindent\textbf{OWOD.} Firstly, we will describe the Open World Object Detection problem in symbolic terms. At time $t$, we use $\mathcal{K}^t = \left \{ 1, 2, ...,C \right \} $ to represent the object classes known to the model. Assume there is a dataset $\mathcal{D}^t = \left \{  \mathcal{X}^t, \mathcal{Y}^t \right \} $ having N images, $ \mathcal{X}^t = \left \{ X_{1} , X_{2} , ..., X_{N} \right \} $ denote the images while $ \mathcal{Y}^t = \left \{  Y_{1} , Y_{2} , ... , Y_{N} \right \} $ denote the corresponding annotations for every image. Specifically, for an  image $X_{i} \in \mathcal{X}^t $, its corresponding $Y_{i} \in \mathcal{Y}^t $ containing $M$ object instance-level annotation $Y_{i} = \left \{  y_{1} , ... , y_{M} \right \}  $, where $M$ is different according to the image contents. Each annotation $y_{m} = \left [ l_{m} , a_{m}, b_{m}, c_{m}, d_{m} \right ]  $ corresponds to one bounding box. Here $l_{m} \in \mathcal{K}^t $ is the category label in known classes and $ \left [ a_{m}, b_{m}, c_{m}, d_{m} \right]  $ is the bounding box coordinates. Since the model is in a dynamic and vast world, there exists unknown object classes denoted as $\mathcal{U}^{t} = \left \{ C+1, ... \right \} $, which may appear during inference time.

In Open World Object Detection, at time $t$, a model trained on $\mathcal{K}^t$ classes is expected to be able to not only identify an instance belonging to any of the known classes $\mathcal{K}^t$, but also recognize an instance from unseen classes $\mathcal{U}^{t}$ by denoting it as 'unknown'. If the annotation information for unseen classes is available, for example, an oracle select $n$ classes from $\mathcal{U}^{t}$ and annotate training labels of them. At time $t+1$,  the known classes set is $ \mathcal{K}^{t+1} =  \left \{1,2,...,C,..,C+n \right \}$. The model can incrementally update itself to detect $C+n$ classes without training from scratch. 

\noindent\textbf{UDA}. The goal of UDA is to classify the samples in $T$ by learning a model with the labeled training samples $\{\mathbf{x}_i,y_i\}_{i=1}^N$ in $S$ and unlabelled ones $\{\mathbf{x}_i\}_{i=1}^M$ in $T$, where $\mathbf{x}_i$ denotes the feature of the $i$-th sample (e.g., an image) extracted by a pre-trained backbone parameterized by $\theta$ (e.g., ResNet-50~\cite{he2016deep} pre-trained on ImageNet~\cite{deng2009imagenet}), and $y_i$ is its ground-truth label. We drop the subscript $i$ for simplicity when the context is clear.
The model includes the backbone, a classification head $f$ and a cluster head $g$, where $f$ and $g$ output the softmax-normalized probability.
Note that $f$ and $g$ have the same output dimension as the classes are shared between $S$ and $T$ in UDA.
UDA's objective is to learn an invariant $f$ that is simultaneously consistent with the classification in $S$, and the clustering in $T$ identified by the cluster head $g$.

\section{Method}
\label{sec:4.1}
\subsection{Motivation}
Object detectors can generate abundant potential object proposals. Only several proposals can be matched with known class ground-truth bounding boxes. Excluding those with large overlaps with ground-truths (e.g., IoU$>$0.5), the rest proposals we refer them as \textbf{unmatched proposals}. These unmatched proposals contain BGs, regions with small overlaps with known class objects, and unknown objects.

State-of-the-art OWOD approaches utilize objectness scores to filter out the unknown objects through the unmatched proposals. Specifically, in ORE~\cite{joseph2021towards}, they directly utilize the attached objectness scores for BG bounding boxes from a head of RPN (Region Proposal Network). OW-DETR~\cite{gupta2022ow} assigns unmatched boxes with their designed multi-scale average activation magnitudes values. After that, both of them select $top\text{-}k$ ones with the highest objectness scores and pseudo-label them as unknown objects, then train the detector.

However, as discussed in Figure~\ref{fig:2}, top-$k$  is a fixed hyper-parameter, \textbf{not flexible}. Given the inherent diversity among images, it is unreasonable to assume that the same top-$k$  can be used across different images to select the same number of proposals and label them all as unknown. Also, it is crucial to note that the objectness scores are derived from detectors that have been trained on known classes. This can introduce a \textbf{bias}: detectors have a tendency to assign higher objectness scores to objects that exhibit similarities with known classes while assigning lower scores to objects that are dissimilar to the known classes. This bias can significantly hinder the detector's performance when it comes to detecting unknown classes. 


Intuitively, we thought of using an unbiased foreground/background (FG/BG for brevity) predictor to replace the top-$k$  proposals selection process. The predictor is a binary classifier, which distinguishes the unmatched proposals into FGs and BGs; then the FGs will be pseudo-labeled as unknown classes. The predictor is supervised by annotated known classes and should mitigate the discrepancy between known class objects and unknown class objects, predicting unknown classes well without annotations.




To build the unbiased predictor, we propose \textbf{P}roposal-\textbf{L}evel \textbf{U}nsupervised Domain Adaptation (PLU) as shown in Figure~\ref{fig:3}. PLU borrows ideas from the UDA (Unsupervised Domain Adaptation), which aims to learn a model in a supervised source domain that generalizes to an unsupervised target domain with a significant domain shift. 



\begin{figure*}
    \centering
    \includegraphics[width = 1.0\linewidth]{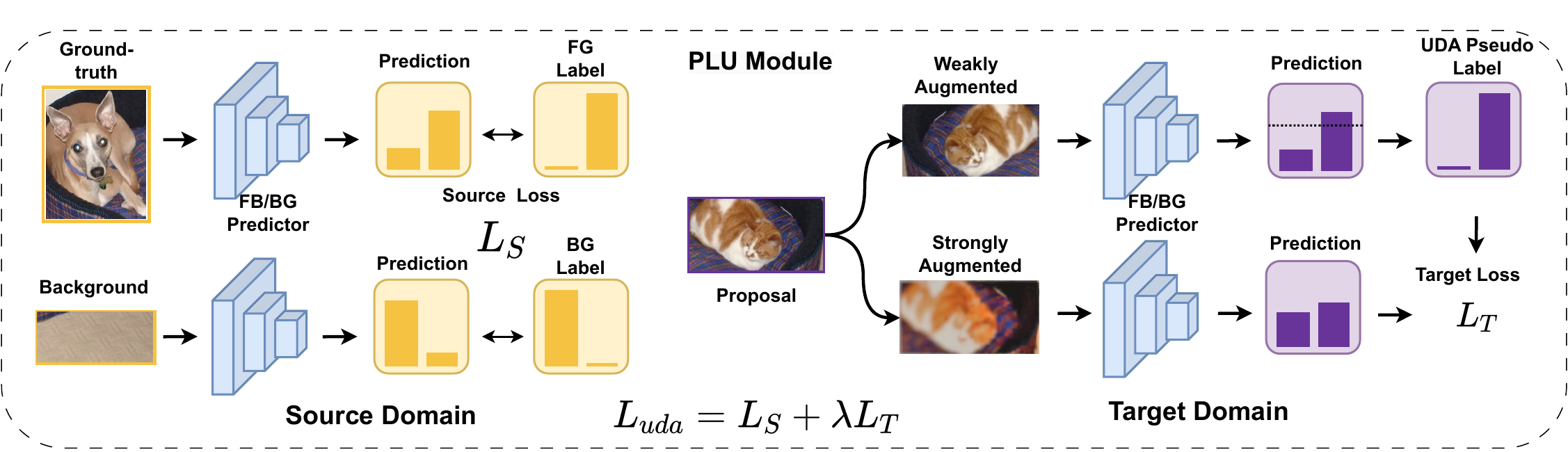}
    \caption{ (a) shows the our overall framework, including the domain formulation. (b) gives more details about the PLU Module.}
    \label{fig:4}
\end{figure*}
\subsection{Domain Formulation}
\label{sec:3.4}

As shown in Figure~\ref{fig:3}a, we use the ground-truths for known classes and unmatched proposals to form the domains:

\noindent\textbf{Source Domain (Known classes)} To transfer the predictor's FG/BG classification ability from the source domain to the target domain, the source domain shall contain FG and BG proposals and their labels. While annotations exist for FG objects in current datasets, there are no corresponding annotations for the BG. As a result, collecting BG samples with high confidence is necessary.

As shown in Figure~\ref{fig:2}c, we empirically find that the proposals thought of as BGs by the detector with very low objectness scores rarely contain objects. Therefore, we collect the BG proposals in online mode: during training, combine the annotated known-class proposal (labeled as `1') and its corresponding image's proposal with the lowest objectness scores (labeled as `0') to be the source domain.

\noindent\textbf{Target Domain (Unknown Classes)} All the unmatched proposals excluding the lowest ones used as Source Domain's BG samples to form the Target Domain. They correspond to locations containing unseen objects or mostly BG but a small part of the seen objects. We do not label them as biased predictions are often ambiguous on them. 

After formulating the Target Domain and Source Domain, we utilize the self-training method~\cite{sakaridis2018model,yang2020fda,zou2018unsupervised,zou2019confidence} in UDA to train an unbiased FG/BG predictor.

\subsection{PLU Module}
Our PLU Module is shown in Figure~\ref{fig:3}b. Here we use FixMatch~\cite{sohn2020fixmatch} as an example of the UDA method, which belongs to a popular class of SSL methods. 


For an input image $X$, we could obtain its Target domain $X_{T}$ from BG proposals that potentially contain an object. Following FixMatch's pipeline, we could weakly augment the Target Domain proposals, obtaining $X_{T_{w}}$ and the strongly-augmented proposals denoted as $X_{T_{s}}$. And we build the input image $X$'s Source domain $X_{S}$ from ground-truth proposals and convincingly BG proposals.  The annotated proposals are labeled as FGs (denote as `1') while the sampled BG proposals that confidently contain no object are labeled as BGs (denote as `0'). Therefore, the FB/BG label $Y_{S_{GT}}$ is a 0-1 vector.


$X_{T_{w}}$, $X_{T_{s}}$, and $X_{S}$ will go through the model denoted as $\phi$. $\phi$ is a binary classification network composed by multi-scale convolutional layers and fully-connected layers to predict whether its input is a FG instance or BG as shown in Eq.~\ref{eq:1}.
\begin{equation}
    \begin{aligned}
        Y_{T_{w}} = \phi(X_{T_{w}}), Y_{T_{s}} = \phi(X_{T_{s}}), Y_{S} = \phi(X_{S})
    \end{aligned}
    \label{eq:1}
\end{equation}

Their output logits are denoted as $Y_{T_{w}}$,$Y_{T_{s}}$,$Y_{S}$.
As FixMatch~\cite{sohn2020fixmatch}'s paradigm, $Y_{T_{w}}$ is used to generate the pseudo label $Y_{T_{p}}$ based on the threshold $\epsilon$, which is a 0-1 vector, as shown in Eq.~\ref{eq:2}.
\begin{equation}
    \centering
    Y_{T_{p}} = \mathop{\max} (softmax(Y_{T_{w}})) > \epsilon
    \label{eq:2}
\end{equation}
\begin{equation}
    \centering
        \begin{aligned}
            L_{T} &= -\left [ Y_{T_{s}} \log_{}{Y_{T_{p}}} + \left ( 1-Y_{T_{s}} \right )  \log_{}{\left ( 1-Y_{T_{p}} \right ) } \right ] \\
            L_{S}  &= -\left [ Y_{S} \log_{}{Y_{S_{GT}}} + \left ( 1-Y_{S} \right )  \log_{}{\left ( 1-Y_{S_{GT}} \right ) } \right ] 
        \end{aligned}
    \label{eq:3}
\end{equation}

Cross-entropy losses are employed to compute loss for $Y_{T_{s}}$ and $Y_{T_{p}}$, $Y_{S}$ and $Y_{S_{GT}}$, as shown in Eq.~\ref{eq:3}. Our full objective for the PLU module is shown in Eq.~\ref{eq:4}, where $\lambda$ controls the relative importance of $L_{T}$ and $L_{S}$.
\begin{equation}
    \centering
    \begin{aligned}
        L_{uda} = L_{T} + \lambda L_{S} 
    \end{aligned}
    \label{eq:4}
\end{equation}

\subsection{General Pipeline of Using PLU}

Note that the PLU module's description does not include any pre-assigned object detection methods and UDA methods. Theoretically, a more widely applicable version of the reformulation pipeline is summarized in Algorithm~\ref{alg:algorithm}.
\begin{algorithm}[!htbp]
\caption{General pipeline of using PLU}
\label{alg:algorithm}
%

\begin{algorithmic}[1] 
\STATE \textbf{Generate enough proposals.} 

The current mainstream object detection framework slides thoroughly on the whole image, and generates enough proposals.\\
\STATE \textbf{Compute objectness (or similar) scores.} 

Assign each proposal with an objectness score. The score represents the possibility the proposal contains an object.\\
\STATE \textbf{Define domains.} 

Define proposals as previously stated. \\
\STATE \textbf{Train.} 

Add PLU module to detection model to train, following a 3-stage training procedure.\\
1) Train a feature extraction backbone. In practice, we usually directly use a self-supervised pre-trained backbone to avoid information leakage.

2) Train the detection model with PLU module, using the UDA loss as the optimization objective.

3) Fine-tune the whole network on a small split of known classes.

\STATE \textbf{Inference}. 
\end{algorithmic}
\label{algorithm 1}
\end{algorithm}


\label{sec:3.3}

\section{Experiment}
\label{sec:4}

\subsection{Implementation Details}

\noindent\textbf{Datasets} We evaluate our model's performance on MS-COCO~\cite{lin2014microsoft}, following the data splits proposed by ORE~\cite{joseph2021towards}. Specifically, the 80 classes MS-COCO images are split into 4 tasks, which can be regarded as 4 times incremental learning. Each task has 20 classes' instances, not overlapping with other tasks, denoted as $\left \{ T_{1}, T_{2}, T_{3}, T_{4} \right \}$. Note that an image can appear in several tasks if it contains instances from multiple tasks.

\noindent\textbf{Network}
For Faster-RCNN-based architecture, we follow ORE's structure, and the IoU threshold for RCNN is 0.5. For DETR-based architecture, we follow the OW-DETR's structure and use 100 queries for every image. We use the ratio of the number of known classes ground-truths to convincing backgrounds as 1:1. We sample the same number of proposals from the unmatched proposals as the image's Source Domain to build its Target Domain. The UDA threshold $\epsilon$ is set to 0.9, and the coefficient $\lambda$ to balance the losses is 1. For the backbone, we use the DINO ResNet-50 pretrained backbone to avoid potential information leakage, which could happen in fully supervised pertaining when there is. For the same concern, we use ResNet-50 training from scratch as the FG/BG predictor in PLU. For each task, we use 4096 samples to train the predictor. We set the batch size to 8. For evaluation,  we use the WI at 0.8. All experiments are conducted on 8 NVIDIA V100 GPUs.



\begin{table*}[t]
\resizebox{1.0\linewidth}{!}{
\renewcommand{\arraystretch}{1.25}
\Large
\begin{tabular}{c|ccc|ccccc|ccccc|ccc}
\cline{1-17}
\cline{1-17}
\textbf{Task IDs} & \multicolumn{3}{c|}{\textbf{Task 1}} & \multicolumn{5}{c|}{\textbf{Task 2}} & \multicolumn{5}{c|}{\textbf{Task 3}} & \multicolumn{3}{c}{\textbf{Task4}}  
\\ \cline{1-17}
 & \multicolumn{2}{c|}{Unknown} & Known (mAP) & \multicolumn{2}{c|}{Unknown} & \multicolumn{3}{c|}{Known (mAP)} & \multicolumn{2}{c}{Unknown} & \multicolumn{3}{c|}{Known (mAP)} & \multicolumn{3}{c}{Known (mAP)} \\ \cline{2-17}
 & WI ($\downarrow$)  & \multicolumn{1}{c|}{U-Recall ($\uparrow$)} & \begin{tabular}[c]{@{}c@{}}Current \\ Known\end{tabular} & WI ($\downarrow$)   & \multicolumn{1}{c|}{U-Recall ($\uparrow$)} & \begin{tabular}[c]{@{}c@{}}Previous \\ Known\end{tabular} & \begin{tabular}[c]{@{}c@{}}Current \\ Known\end{tabular} & Both & WI ($\downarrow$) & U-Recall ($\uparrow$) & \begin{tabular}[c]{@{}c@{}}Previous \\ Known\end{tabular} & \begin{tabular}[c]{@{}c@{}}Current \\ Known\end{tabular} & Both & \begin{tabular}[c]{@{}c@{}}Previous \\ Known\end{tabular} & \begin{tabular}[c]{@{}c@{}}Current \\ Known\end{tabular} & Both \\ \cline{1-17}
 Faster-RCNN & \textbf{--} & \multicolumn{1}{c|}{\textbf{--}} & 56.4 & \textbf{--} & \multicolumn{1}{c|}{\textbf{--}} & 3.7 & 26.7 & 15.2 & \textbf{--} & \textbf{--} & 2.5 & 15.2 & 6.7 & 0.8 & 14.5 & 4.2  \\
ORE$-$EBUI & 0.048 & \multicolumn{1}{c|}{6.9} & 56.1 & 0.029 & \multicolumn{1}{c|}{4.2} & 51.9 & 26.2 & 39.1 & 0.020 & 5.1 & 39.2 & 13.3 & 30.6 & 30.3 & 12.8 & 25.9 \\
Faster-RCNN+PLU & \textbf{0.045} & \multicolumn{1}{c|}{\textbf{9.2}} & \textbf{57.6} & \textbf{0.027}  & \multicolumn{1}{c|}{3.8} & \textbf{52.5} & \textbf{31.2} & \textbf{41.8} & \textbf{0.018} & \textbf{5.8} & \textbf{42.1} & \textbf{17.8} & \textbf{34.0} & \textbf{32.6} & \textbf{15.4} & \textbf{28.3} \\ \cline{1-17}
DDETR & \textbf{--} & \multicolumn{1}{c|}{\textbf{--}} & 60.3 & \textbf{--} & \multicolumn{1}{c|}{\textbf{--}} & 4.5 & 31.3 & 17.9 & \textbf{--} & \textbf{--} & 3.3 & 22.5 & 8.5 & 2.5 & 16.4 & 6.2  \\
OW-DETR & 0.051 & \multicolumn{1}{c|}{7.0} & 58.9 & 0.035 & \multicolumn{1}{c|}{5.5} & 53.1 & 32.9 & 42.7 & 0.026 & 6.0 & 38.5 & 14.2 & 30.2 & 31.0 & 12.7 & 26.3 \\
DETR+PLU & 
\textbf{0.034} & \multicolumn{1}{c|}{\textbf{10.5}} & \textbf{61.4} & \textbf{0.029} & \multicolumn{1}{c|}{\textbf{7.4}} & \textbf{55.8} &  \textbf{35.6} & \textbf{45.7} & \textbf{0.020} & \textbf{6.6} & \textbf{41.8} & \textbf{18.9} & \textbf{34.2} & \textbf{34.1} & \textbf{16.2} & \textbf{29.6} \\ \cline{1-17}

\cline{1-17}
\cline{1-17}
\end{tabular}
}
{\vspace{0.2em}}
\caption{State of the art comparison for OWOD on MS-COCO for unknown classes. This comparison shows the comparison of current OWOD approaches without/with our PLU module based on Faster-RCNN and structures respectively. }

\label{tab:2}
\end{table*}

\begin{table*}[t]
    \begin{subtable}[h]{0.4\textwidth}
        \centering
        \begin{tabular}{l|cc|ccc}
        \hline \hline
        FG/BG & WI & U-Recall & {Previous} & {Current} & Both \\ \hline
        1:1 & \cellcolor[HTML]{ECEDFF}0.027 & \cellcolor[HTML]{ECEDFF}3.8 & \cellcolor[HTML]{ECEDFF}52.5  & \cellcolor[HTML]{ECEDFF}31.2 & \cellcolor[HTML]{ECEDFF}41.8 \\ 
        1:2 & 0.031 & 3.2  & 51.6 & 28.7 & 40.2 \\ 
        1:5 & 0.089 & 1.5  & 46.7 & 26.3 & 36.5 \\
        1:10 & 0.133 & 0.4  & 43.3 & 24.5 & 33.9 \\ 
        \hline \hline
        \end{tabular}
       \caption{Ablation on FG/BG sample ratio.}
       \label{tab:week1}
    \end{subtable}
    \hfill
    \begin{subtable}[h]         
        {0.5\textwidth}
        \centering
        \begin{tabular}{c|cc|ccc}
        \hline \hline
        $\lambda$ & WI & U-Recall & {Previous} & {Current} & Both \\ \hline
        1& \cellcolor[HTML]{ECEDFF}0.027 & \cellcolor[HTML]{ECEDFF}3.8 & \cellcolor[HTML]{ECEDFF}52.5  & \cellcolor[HTML]{ECEDFF}31.2 & \cellcolor[HTML]{ECEDFF}41.8 \\ 
        0.7 & 0.055 & 2.4  & 48.3 & 27.4 & 37.9 \\ 
        0.5 & 0.104 & 0.8  & 44.2 & 25.1 & 34.7 \\
        0.2 & 0.221 & 0.2  & 39.8 & 17.6 & 28.7 \\ 
        \hline \hline
        \end{tabular}
        \caption{Ablation on $\lambda$.}
        \label{tab:week2}
     \end{subtable}
     \\
     \begin{subtable}[h]{0.4\textwidth}
        \centering
        \begin{tabular}{l|cc|ccc}
        \hline \hline
        Method & WI & U-R & {Previous} & {Current} & Both \\ \hline
        F-R+FT & \cellcolor[HTML]{ECEDFF}0.027 & \cellcolor[HTML]{ECEDFF}3.8 & \cellcolor[HTML]{ECEDFF}52.5  & \cellcolor[HTML]{ECEDFF}31.2 & \cellcolor[HTML]{ECEDFF}41.8 \\ 
        F-R w/o FT & 0.065 & 1.3 & 11.8 & 19.1 & 15.5 \\ 
        DTR + FT & \cellcolor[HTML]{ECEDFF}0.032 & \cellcolor[HTML]{ECEDFF}6.2  & \cellcolor[HTML]{ECEDFF}55.8 & \cellcolor[HTML]{ECEDFF}35.6 & \cellcolor[HTML]{ECEDFF}45.7 \\
        DTR w/o FT & 0.031 & 6.8 & 9.4 & 25.0 & 17.2\\ 
        \hline \hline
        \end{tabular}
        \caption{Ablation on w/o fine-tuning }
        \label{tab:week2}
     \end{subtable}
     \hfill
    \begin{subtable}[h]         
        {0.5\textwidth}
        \centering
        \begin{tabular}{l|cc|ccc}
        \hline \hline
        Method & WI & U-R & {Previous} & {Current} & Both \\ \hline
        F-R+CST & \cellcolor[HTML]{ECEDFF}0.027 & \cellcolor[HTML]{ECEDFF}3.8 & \cellcolor[HTML]{ECEDFF}52.5  & \cellcolor[HTML]{ECEDFF}31.2 & \cellcolor[HTML]{ECEDFF}41.8 \\ 
        F-R+FM & 0.031 & 3.2  & 52.3 & 30.3 & 41.3 \\ 
        DTR+CST & \cellcolor[HTML]{ECEDFF}0.029 & \cellcolor[HTML]{ECEDFF}7.4  & \cellcolor[HTML]{ECEDFF}55.8 & \cellcolor[HTML]{ECEDFF}35.6 & \cellcolor[HTML]{ECEDFF}45.7 \\
        DTR+FM & 0.035 & 6.0  & 55.1 & 34.5 & 44.8 \\ 
        \hline \hline
        \end{tabular}
        \caption{Ablation on different UDA methods.}
        \label{tab:week2}
     \end{subtable}
     \caption{Ablation results. Models are trained on Task 2. Our default settings are in lavender. (U-R: U-Recall, F-R: Faster-RCNN+PLU, FT: Finetuning, DTR: DETR+PLU, FM: FixMatch)}
     \label{tab:ablation}
\end{table*}
\noindent\textbf{Evaluation Metrics}
We use the common mean average precision (mAP) at 0.5 IOU threshold as the metric for previous known and current known classes. 

As for unknown classes, we use WI, A-OSE and U-Recall as metrics following previous literature~\cite{joseph2021towards,gupta2022ow}. The WI (Wilderness Impact~\cite{dhamija2020overlooked}) metric is computed by the model's precision evaluated on known classes.

$P_{\mathcal{K}}$ and precision on known and unknown classes $P_{\mathcal{K} \cup \mathcal{U}}$. The model with a smaller WI has better performance in distinguishing between known and unknown classes.
\begin{equation}
    WI=\frac{P_{\mathcal{K}}}{P_{\mathcal{K} \cup  \mathcal{U}}}-1
\end{equation}


The U-Recall (Unknown Recall) metric is the recall for unknown classes, showing the model's performance to detect unknown objects exhaustively.

\subsection{Main Results}
\label{sec:5.1}
Table~\ref{tab:2} shows the comparison on the open-world evaluation protocol. To ensure fairness, we perform a comparison of our PLU model against other methods using Faster-RCNN-based and DETR-based structures separately.

For Faster-RCNN-based approaches, we report our method, vanilla Faster-RCNN and ORE~\cite{joseph2021towards} without the energy-based~\cite{lecun2006tutorial} unknown identifier (EBUI), which is used for inference based on a Weibull distribution fitting on held-out validation data.

For DETR-based approaches, we present vanilla DDETR~\cite{zhu2020deformable}, OW-DETR~\cite{gupta2022ow}, and our PLU performance. For fairness, we do not compare with multi-modal transformers~\cite{kamath2021mdetr,maaz2022class} which uses text as another modal and uses image-text pairs pretrained backbone, which might result in information leakage.  

We compare their performance on known classes in terms of mAP, unknown classes in terms of WI, and U-Recall. For Task 1, there are no previous known classes, so previous known mAP is not computed. As for Task 4, all annotated classes have been introduced. There are no unknown instances in the dataset so the WI and U-Recall scores are not applicable. 

The results show that PLU has trained a FG/BG predictor, which improves the current OWOD detector performance on the retrieval of unknown objects, leading to improved performance with significant gains for WI, U-Recall, on the same tasks 1, 2, and 3. Furthermore, PLU outperforms the best existing Faster-RCNN based OWOD approach of ORE in terms of the known class mAP on all four tasks. As classes are gradually introduced, the performance gains has been cumulatively larger up to $4.6\%$. A similar trend is also shown in Transformer-based approaches. Known classes have $0.5\sim2.2$ absolute performance increase while unknown performance is also better. 





\begin{figure*}[!htbp]
  \centering
  {\includegraphics[width=1.0\linewidth]{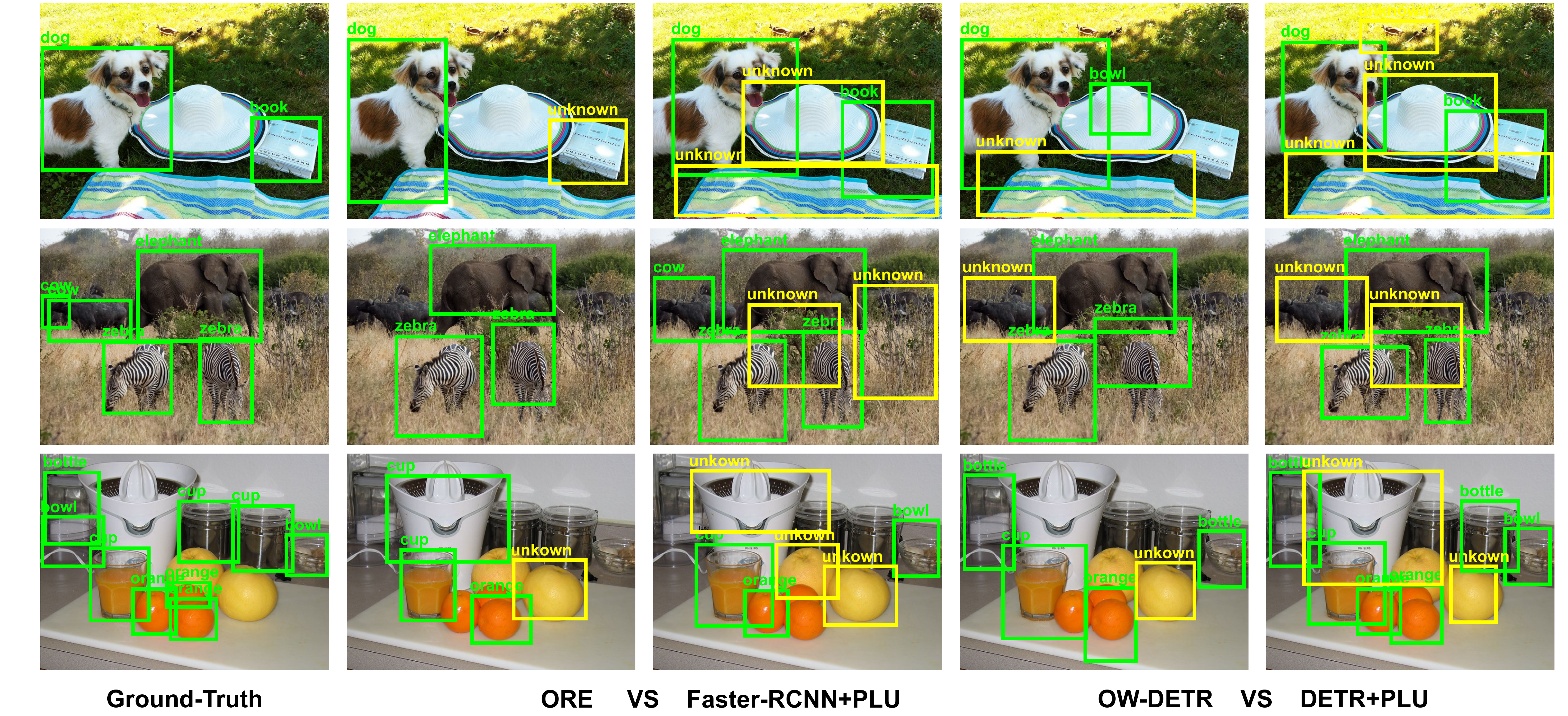}}
   \caption{Qualitative Results. We show the ground-truth label, the Faster-RCNN-based prediction comparison, and the DETR-based prediction comparison, respectively. \textcolor{green}{Green} is for known, and \textcolor{yellow}{yellow} is for unknown.}
   \label{fig:6}
\end{figure*}
\begin{table*}[t]
\resizebox{1\linewidth}{!}{
    \renewcommand{\arraystretch}{1.25}
    \Large
    \begin{tabular}{c|ccccccccccccccccccccc}
    \hline
    \textcolor{blue}{\textbf{10+10 setting}} & \multicolumn{1}{c}{aero} & \multicolumn{1}{c}{cycle} & bird & boat & \multicolumn{1}{c}{bottle} & bus & car & \multicolumn{1}{c}{cat} & chair & cow & table & dog & horse & bike & person & plant & sheep & sofa & train & tv & mAP \\ \hline
    ILOD~\cite{shmelkov2017incremental} & 69.9 & 70.4 & 69.4 & 54.3 & 48.0 & 68.7 & 78.9 & 68.4 & 45.5 & 58.1 & \cellcolor[HTML]{ECEDFF}59.7 & \cellcolor[HTML]{ECEDFF}72.7 & \cellcolor[HTML]{ECEDFF} 73.5 & \cellcolor[HTML]{ECEDFF}73.2 & \cellcolor[HTML]{ECEDFF}66.3 & \cellcolor[HTML]{ECEDFF}29.5 & \cellcolor[HTML]{ECEDFF}63.4 & \cellcolor[HTML]{ECEDFF}61.6 & \cellcolor[HTML]{ECEDFF}69.3 & \cellcolor[HTML]{ECEDFF}62.2 & 63.2 \\ 
    ORE$-$EBUI~\cite{joseph2021towards} & 63.5 & 70.9 & 58.9 & 42.9 & 34.1 & 76.2 & 80.7 & 76.3 & 34.1 & 66.1 & \cellcolor[HTML]{ECEDFF}56.1 & \cellcolor[HTML]{ECEDFF}70.4 & \cellcolor[HTML]{ECEDFF}80.2 & \cellcolor[HTML]{ECEDFF}72.3 & \cellcolor[HTML]{ECEDFF}81.8 & \cellcolor[HTML]{ECEDFF}42.7 & \cellcolor[HTML]{ECEDFF}71.6 & \cellcolor[HTML]{ECEDFF}68.1 & \cellcolor[HTML]{ECEDFF}77.0 & \cellcolor[HTML]{ECEDFF}67.7 & 64.5 \\ 
    Faster-RCNN+PLU & 66.9 & 66.8 & 63.1 & 53.0 & 45.4 & 75.1 & 82.7 & 73.1 & 36.0 & 67.2 & 
    \cellcolor[HTML]{ECEDFF}57.3 & \cellcolor[HTML]{ECEDFF}71.4 & \cellcolor[HTML]{ECEDFF}80.6 & \cellcolor[HTML]{ECEDFF}74.7 & \cellcolor[HTML]{ECEDFF}75.8 & \cellcolor[HTML]{ECEDFF}38.9 & \cellcolor[HTML]{ECEDFF}69.2 & \cellcolor[HTML]{ECEDFF}67.3 & \cellcolor[HTML]{ECEDFF}69.8 & \cellcolor[HTML]{ECEDFF}64.0 & \textbf{64.9} 
    \\ \cline{1-22}
    OW-DETR~\cite{gupta2022ow} & 61.8 & 69.1 & 67.8 & 45.8 & 47.3 & 78.3 & 78.4 & 78.6 & 36.2 & 71.5 & 
    \cellcolor[HTML]{ECEDFF}57.5 & \cellcolor[HTML]{ECEDFF}75.3 & \cellcolor[HTML]{ECEDFF}76.2 & \cellcolor[HTML]{ECEDFF}77.4 & \cellcolor[HTML]{ECEDFF}79.5 & \cellcolor[HTML]{ECEDFF}40.1 & \cellcolor[HTML]{ECEDFF}66.8 & \cellcolor[HTML]{ECEDFF}66.3 & \cellcolor[HTML]{ECEDFF}75.6 & \cellcolor[HTML]{ECEDFF}64.1 & 65.7 \\ 
    DETR+PLU& 70.4 & 67.3 & 64.9 & 56.3 & 52.9 & 79.5 & 80.4 & 77.5 & 39.2 & 74.8 & 
    \cellcolor[HTML]{ECEDFF}56.9 & \cellcolor[HTML]{ECEDFF}73.4 & \cellcolor[HTML]{ECEDFF}69.7 & \cellcolor[HTML]{ECEDFF}77.1 & \cellcolor[HTML]{ECEDFF}80.9 & \cellcolor[HTML]{ECEDFF}40.2 & \cellcolor[HTML]{ECEDFF}70.0 & \cellcolor[HTML]{ECEDFF}72.6 & \cellcolor[HTML]{ECEDFF}75.3 & \cellcolor[HTML]{ECEDFF}59.9 & \textbf{66.9} \\\hline \hline
    \textcolor{blue}{\textbf{15+5 setting}} & \multicolumn{1}{c}{aero} & \multicolumn{1}{c}{cycle} & bird & boat & \multicolumn{1}{c}{bottle} & bus & car & \multicolumn{1}{c}{cat} & chair & cow & table & dog & horse & bike & person & plant & sheep & sofa & train & tv & mAP \\ \hline
    ILOD~\cite{shmelkov2017incremental} & 70.5 & 79.2 & 68.8 & 59.1 & 53.2 & 75.4 & 79.4 & 78.8 & 46.6 & 59.4 & 59.0 & 75.8 & 71.8 & 78.6 & 69.6 & \cellcolor[HTML]{ECEDFF}33.7 & \cellcolor[HTML]{ECEDFF}61.5 & \cellcolor[HTML]{ECEDFF}63.1 & \cellcolor[HTML]{ECEDFF}71.7 & \cellcolor[HTML]{ECEDFF}62.2 &  65.8 \\ 
    ORE$-$EBUI~\cite{joseph2021towards} & 75.4 & 81.0 & 67.1  & 51.9 & 55.7 & 77.2 & 85.6 & 81.7 & 46.1 & 76.2 & 55.4 & 76.7 & 86.2 & 78.5 & 82.1 & \cellcolor[HTML]{ECEDFF}32.8 & \cellcolor[HTML]{ECEDFF}63.6 & \cellcolor[HTML]{ECEDFF}54.7 & \cellcolor[HTML]{ECEDFF}77.7 & \cellcolor[HTML]{ECEDFF}64.6 & 68.5 \\ 
    Faster-RCNN+PLU & 76.4 & 79.2 & 80.6 & 60.8 & 53.6 & 70.2 & 85.4 & 84.6 & 43.4 & 74.0 & 57.1 & 80.4 & 85.2 & 78.9 & 84.2 & 
    \cellcolor[HTML]{ECEDFF}29.6 & \cellcolor[HTML]{ECEDFF}61.9 & \cellcolor[HTML]{ECEDFF}49.6 & \cellcolor[HTML]{ECEDFF}75.5 & \cellcolor[HTML]{ECEDFF}62.4 & \textbf{68.7} \\ \cline{1-22}
    OW-DETR~\cite{gupta2022ow} & 77.1 & 76.5 & 69.2 & 51.3 & 61.3 & 79.8 & 84.2 & 81.0 & 49.7 & 79.6 & 58.1 & 79.0 & 83.1 & 67.8 & 85.4 & 
    \cellcolor[HTML]{ECEDFF}33.2 & 
    \cellcolor[HTML]{ECEDFF}65.1 & \cellcolor[HTML]{ECEDFF}62.0 & 
    \cellcolor[HTML]{ECEDFF}73.9 & 
    \cellcolor[HTML]{ECEDFF}65.0 & \textbf{69.4} \\ 
    DETR+PLU & 78.6 & 77.4 & 63.4 & 57.2 & 59.2 & 72.7 & 79.9 & 85.4 & 47.0 & 76.1 & 61.6 & 79.7 & 85.1 & 68.5 & 79.4 & 
    \cellcolor[HTML]{ECEDFF}34.8 & \cellcolor[HTML]{ECEDFF}70.8 & \cellcolor[HTML]{ECEDFF}58.0 & \cellcolor[HTML]{ECEDFF}82.4 & \cellcolor[HTML]{ECEDFF}64.5 & 69.1\\ \hline \hline
    \textcolor{blue}{\textbf{19+1 setting}} & \multicolumn{1}{c}{aero} & \multicolumn{1}{c}{cycle} & bird & boat & \multicolumn{1}{c}{bottle} & bus & car & \multicolumn{1}{c}{cat} & chair & cow & table & dog & horse & bike & person & plant & sheep & sofa & train & tv & mAP \\ \hline
    ILOD~\cite{shmelkov2017incremental} & 69.4 & 79.3 & 69.5 & 57.4 & 45.4 & 78.4 & 79.1 & 80.5 & 45.7 & 76.3 & 64.8 & 77.2 & 80.8 & 77.5 & 70.1 & 42.3 & 67.5 & 64.4 & 76.7 & \cellcolor[HTML]{ECEDFF}62.7 & 68.2 \\ 
    ORE$-$EBUI~\cite{joseph2021towards} 
    & 67.3 & 76.8 & 60.0 & 48.4 & 58.8 & 81.1 & 86.5 & 75.8 & 41.5 & 79.6 & 54.6 & 72.8 & 85.9 & 81.7 & 82.4 & 44.8 & 75.8 & 68.2 & 75.7 & \cellcolor[HTML]{ECEDFF}60.1 & 68.8 \\ 
    
    Faster-RCNN+PLU 
    & 74.7 & 78.4 & 65.8 & 47.7 & 55.7 & 75.1 & 85.4 & 84.9 & 42.3 & 81.5 & 57.5 & 78.8 & 84.6 & 80.7 & 82.2  & 37.1 & 75.1 & 63.5 & 71.6 & \cellcolor[HTML]{ECEDFF}59.4 & \textbf{69.1}
    
    \\ \hline
    OW-DETR~\cite{gupta2022ow} & 70.5 & 77.2 & 73.8 & 54.0 & 55.6 & 79.0 & 80.8 & 80.6 & 43.2 & 80.4 & 53.5 & 77.5 & 89.5 & 82.0 & 74.7 & 43.3 & 71.9 & 66.6 & 79.4 & \cellcolor[HTML]{ECEDFF}62.0 & 70.2 \\ 
    DETR+PLU 
    & 76.2 & 81.4 & 71.2 & 51.8 & 54.6 & 77.4 & 84.7 & 85.9 & 47.0 & 83.1 & 60.1 & 82.0 & 85.6 & 81.5 & 82.2 & 42.3 & 75.3 & 65.1 & 78.1& \cellcolor[HTML]{ECEDFF}60.5 & \textbf{71.3}\\ \hline
\end{tabular}
}
{\vspace{0.2em}}
\caption{Comparison for incremental object detection (iOD) on PASCAL VOC. Unknown classes are in lavender. }
\label{tab:3}
\end{table*}
\begin{table}[!htbp]
   \centering
  \resizebox{1\linewidth}{!}{
    \renewcommand{\arraystretch}{1.25}
    \Large
    \begin{tabular}{c|cc}
    \cline{1-3}
        Evaluated on & VOC 2007 & VOC 2007 + COCO (WR1) \\ \cline{1-3}
        Standard RetinaNet & 79.2 & 73.8 \\
        Dropout Sampling & 78.1 & 71.1  \\ 
        ORE & 80.2 & 77.9   \\
        OR +PLU & \textbf{81.1} & \textbf{78.3} \\ 
        \cline{1-3}
        OW-DETR & 81.4 & 77.6   \\
        OW-DETR+PLU & \textbf{81.9} & \textbf{78.5} \\
        \cline{1-3}
  \end{tabular}}
  \caption{Open-set object detection
comparison}
  \label{tab:4}
\end{table}
\subsection{Ablation Study}
\label{sec:4.4}
\noindent \textbf{FG/BG Samples Ratio in Source Domain} We change the ratios of FG proposals (ground-truth proposals) to BG proposals sampled from each image, which builds the Source Domain. Its ablation results on Faster-RCNN-based framework are shown in Table~\ref{tab:ablation}a. We use 1:1 in all experiments.


\noindent \textbf{Choice of $\lambda$} $\lambda$ controls the relative importance of $L_{T}$ and $L_{S}$. Its ablation results on Faster-RCNN-based framework in Table~\ref{tab:ablation}b. We adopt $\lambda = 1$ in our all experiments.

\noindent \textbf{w/o fine-tuning on known classes} As our general pipeline~\ref{algorithm 1} mentioned, we need to fine-tune the whole network on a small split of known classes, which helps counter catastrophic forgetting. We ablate the fine-tuning process in Table~\ref{tab:ablation}c. 

\noindent \textbf{Different UDA methods} As previously mentioned, different kinds of UDA methods can adapt to our PLU pipeline. In Table~\ref{tab:ablation}d, we use two UDA methods, FixMatch and CST to show that different UDA methods can work well. 

\subsection{Qualitative Results}

When comparing the Faster-RCNN-based methods, as shown in Figure~\ref{fig:6}'s second and third columns, our method could not only detect the known class more precisely than ORE but also detect more unknown objects such as the hat, and the tree behind zebras. Additionally, the kettle misclassified by ORE is correctly labeled `unknown' by our method. 

When comparing the DETR-based methods, as shown in Figure~\ref{fig:6}'s fourth and fifth columns, our method correctly detects the hat as `unknown' while OW-DETR fails. For occluded unknown objects, our method performs better \eg find the tree behind zebras and the electric juice press.

\subsection{By-product Experiment}
\label{appendix: by-product experiments}
\noindent\textbf{Incremental Object Detection }
As shown in Table~\ref{tab:3}, the PLU module has demonstrated improved performance on incremental object detection (iOD) tasks as a byproduct of its enhanced ability to detect unknown objects. Table~\ref{tab:3} presents the results of experiments conducted on three different iOD tasks using PASCAL VOC 2007 dataset. These tasks involved the introduction of 10, 15, and 19 classes, followed by the incremental introduction of the remaining 10, 5, and 1 classes, respectively, and the evaluation of their performance on an evaluation protocol for 20 classes. The performance of Faster-RCNN based detectors with our PLU modules is better than existing approaches in the incremental object detection settings. Similar trends can be observed in the DETR-based results.

\noindent\textbf{Open-set Performance} As Table~\ref{tab:4} shows, the mAP values obtained by evaluating the detector on closed set data (trained and tested on Pascal VOC 2007) as well as on open-set data (test set containing an equal number of unknown images from MS-COCO) provide a reliable metric for assessing the detector's ability to handle unknown instances. We follow the protocol in ORE and we find out that PLU is better at handling the performance drop.

\section{Conclusion}
\label{sec:5}
In this paper, a new approach called PLU is proposed to address the limitations of the previous OWOD method's top-$k$ selection strategy. PLU employs a predictor that is trained under a self-training UDA approach to obtain discriminative features for distinguishing unknown objects and backgrounds. The PLU pipeline is compatible with mainstream detection models and UDA methods. The proposed approach is evaluated on two frameworks, Faster-RCNN and DETR, and achieves state-of-the-art performance.


{\small
\bibliographystyle{ieee_fullname}
\bibliography{references}
}

\end{document}